\documentclass{article}

\usepackage[main,preprint]{neurips_2026}

\usepackage[utf8]{inputenc} 
\usepackage[T1]{fontenc}    
\usepackage{url}            
\usepackage{booktabs}       
\usepackage{amsfonts}       
\usepackage{nicefrac}       
\usepackage{microtype}      
\usepackage{xcolor}         

\usepackage[utf8]{inputenc} 
\usepackage[T1]{fontenc}    
\usepackage{url}            
\usepackage{booktabs}       
\usepackage{amsfonts}       
\usepackage{nicefrac}       
\usepackage{microtype}      
\usepackage{xcolor}         

\definecolor{mydarkblue}{rgb}{0,0.08,0.45}
\usepackage[
colorlinks=true,
linkcolor=mydarkblue,
citecolor=mydarkblue,
filecolor=mydarkblue,
urlcolor=mydarkblue,
]{hyperref}

\usepackage{amsmath}
\usepackage{amssymb}
\usepackage{mathtools}
\usepackage{amsthm}

\usepackage{multirow}
\usepackage{multicol}
\usepackage{enumitem}
\usepackage{tcolorbox}

\newtheorem{obs}{Observation}
\NewDocumentEnvironment{observation}{+b}
  {
    \begin{tcolorbox}[colback=gray!10, colframe=black, boxrule=1pt, arc=4pt, left=5pt, right=5pt]
    \begin{obs}
      #1
    \end{obs}
    \end{tcolorbox}
  }{}

\renewcommand{\paragraph}[1]{%
  \noindent\textbf{#1}\hspace{1em}\ignorespaces%
}

\title{Turning Tabular Foundation Models \\ into Graph Foundation Models}

\author{%
Dmitry Eremeev\thanks{Correspondence to: \texttt{eremeev-d@yandex-team.ru}} \\
HSE University \\
Yandex Research \\
\And
Gleb Bazhenov \\
HSE University \\
Yandex Research \\
\And
Oleg Platonov \\
HSE University \\
Yandex Research \\
\AND
Artem Babenko \\
Yandex Research \\
\And
Liudmila Prokhorenkova \\
Yandex Research \\
}

\begin{document}

\maketitle

\begin{abstract}
While foundation models have revolutionized fields such as natural language processing and computer vision, their potential in graph machine learning remains largely unexplored. One of the key challenges in designing graph foundation models (GFMs) is handling diverse node features that can vary across different graph datasets. While many works on GFMs have focused exclusively on text-attributed graphs, the problem of handling arbitrary features of other types in GFMs has not been fully addressed. However, this problem is not unique to the graph domain, as it also arises in the field of machine learning for tabular data. In this work, motivated by the recent success of tabular foundation models (TFMs) like TabPFNv2 and LimiX, we propose G2T-FM, a simple framework that allows tabular foundation models to be applied to graph node-level tasks. Specifically, G2T-FM augments the original node features with neighborhood feature aggregation, adds structural embeddings, and then applies a TFM to the constructed node representations. Even in the in-context learning setting, our model achieves strong results when combined with a strong TFM, outperforming both prior GFMs and well-tuned GNNs trained from scratch. Moreover, after finetuning, G2T-FM consistently surpasses well-tuned GNN baselines, often by a significant margin. In summary, our paper reveals the potential of a previously overlooked direction: utilizing tabular foundation models for graph machine learning tasks. Our source code is available at \url{https://github.com/yandex-research/g2t-fm}.
\end{abstract}

\section{Introduction}

In recent years, foundation models have become a major breakthrough in deep learning. Foundation models are large machine learning models that are pretrained on diverse and extensive datasets. After this pretraining phase, they can be easily adapted to a variety of specific tasks with minimal additional training. Well-known examples include BERT~\citep{BERT} and GPT~\citep{brown2020language} in natural language processing, as well as CLIP~\citep{CLIP} in computer vision. The core principle behind foundation models is to learn general representations by leveraging large and varied data. These representations capture important patterns and semantics in the data, making the pretrained models highly transferable to different tasks. As a result, foundation models consistently achieve state-of-the-art results, while also improving efficiency and generalization. Furthermore, these models unify techniques across different fields, driving rapid progress and innovation in deep learning research and applications.

Despite their remarkable success in areas such as computer vision and natural language processing, the development of foundation models for graph data has not advanced to the same extent. The challenges of developing  graph foundation models (GFMs) stem from the fact that graphs are not actually a single domain, but rather a way to represent data from different domains. These domains use graphs to represent very different structures, such as social networks, web networks, road networks, co-purchasing networks, molecules, connectomes, or even abstract objects and their relations. 
Thus, successful GFMs should be able to work with graphs from different domains where nodes represent very different objects and edges represent very different relations, which is a rather formidable task that requires overcoming many serious challenges. 
Two key challenges faced by GFMs are the ability to transfer to new feature spaces and target spaces. Graphs from different domains often have different node features and different targets, making it difficult to design GFMs that can work across various types of graphs. Some existing GFMs restrict themselves to text-attributed graphs~\citep{GFT, UniGraph, OFA}, which allows them to use pretrained text encoders. Another approach is to use simple dimensionality reduction methods like SVD and PCA~\citep{AnyGraph, GCOPE, MDGFM, SAMGPT}, which allow transforming all feature spaces to a space with a fixed predefined number of features. However, these approaches do not allow for the full and effective leveraging of arbitrary node features in graphs from new domains.

The challenges of transferring to new feature and target spaces are not, however, exclusive to graphs. Tabular data~--- one of the most widespread data modalities in machine learning~--- is similar to graph-structured data in that it does not constitute a single domain but is a way to represent data from different domains. Thus, tabular datasets come with different feature and target spaces, so tabular foundation models (TFMs) face issues similar to those faced by GFMs. While TFMs are not as developed as foundation models for language or vision, they have seen increased interest recently~\citep{breugel2024tabular}, with the first successful approaches being proposed~\citep{TabPFN, TabPFNv2, mueller2025mothernet, ma2024tabdpt, qu2025tabicl}. For instance, TabPFNv2~\citep{TabPFNv2} demonstrates strong performance in both in-context and finetuning settings, and it has recently gained significant attention from the community.

This parallel suggests that developers of GFMs can draw inspiration from TFMs, which have to deal with many of the same challenges. In this paper, we take a first step in this direction and show that tabular foundation models, such as TabPFNv2~\citep{TabPFNv2} and LimiX~\citep{LimiX}, can be effectively adapted to graph datasets. We introduce a simple framework named \underline{G}raph-to-\underline{T}able \underline{F}oundation \underline{M}odel (G2T-FM), which transforms graph node-level tasks into tabular ones and solves them with a tabular foundation model. More specifically, we augment the original features with neighborhood feature aggregations~\citep{GraphLand}, classical structure-based features (node degree, PageRank, and the eigenvectors of the graph Laplacian), and learnable structure-based encodings~\citep{PEARL}. Then, we apply a tabular foundation model to the constructed node representations to get predictions. 

Our empirical results indicate that this straightforward framework achieves strong results in a fully in-context learning setting, outperforming both prior GFMs and well-tuned GNNs trained from scratch. Moreover, after finetuning, G2T-FM surpasses well-tuned GNN baselines, with especially strong improvements obtained when G2T-FM uses LimiX as a tabular foundation model (see Table~\ref{tab:graphland}). These results highlight the potential of the proposed approach and the positive transfer brought by the usage of foundation models. 

Importantly, G2T-FM can be directly applied to any existing or future TFM. Given the rapid pace of development of tabular foundation models~\citep{TabPFNv2_5, qu2026tabiclv2}, in terms of both downstream quality and scalability, G2T-FM can directly benefit from it, and thus we believe that G2T-FM will remain an important and up-to-date baseline even with further development of GFMs.

Our main contributions can be summarized as follows.
\begin{itemize}[leftmargin=10pt, topsep=1pt, beginpenalty=10000]
\setlength\itemsep{1pt}
\item We identify a promising and previously overlooked direction of applying tabular foundation models to graph machine learning tasks.
\item As a proof of concept, we introduce G2T-FM, a simple framework that transforms graph node-level tasks into tabular ones and solves them with a tabular foundation model.
\item We show that G2T-FM serves as a strong baseline for GFMs, outperforming both publicly available GFMs and well-tuned traditional GNNs even in the in-context learning setting, and that finetuning further improves performance.
\end{itemize}

\section{Background}

\subsection{Graph foundation models for node classification}

Graph foundation models (GFMs) have recently gained significant attention in the field of graph machine learning. The main purpose of GFMs is to enable effective transfer of knowledge across different graph datasets. In other words, they aim to learn knowledge from a variety of graph tasks that can be successfully applied to other graphs.
In this work, we primarily focus on node-level tasks such as node classification and node regression. While many GFMs are limited to text-attributed graphs \citep{GFT, UniGraph, OFA}, graphs in many domains involve non-textual features. Therefore, we specifically discuss methods applicable to graphs with arbitrary numerical and categorical node features, which frequently appear in various real-world industrial applications.

In the following sections, we review the key design choices and considerations in the development of GFMs. In particular, we focus on pretraining objectives and data, as well as how GFMs handle graph structure and node features. For further details, see the survey by \citet{wang2025graph}.

\paragraph{Pretraining objective} Some graph foundation models use self-supervised learning (SSL) objectives to guide their pretraining process \citep{GCOPE, SAMGPT, MDGFM}, whereas others employ supervised learning strategies \citep{GraphFM, TSGNN}. Notably, several works \citep{OpenGraph, AnyGraph} reduce the node classification task to link prediction. Specifically, for each label in a downstream task, they create a virtual node that is connected to all training nodes of that class. Node classification thus becomes a task of predicting links to those virtual class nodes, for which the models are pretrained.

\paragraph{Pretraining data} Collecting a sufficiently diverse collection of datasets for pretraining GFMs remains a significant challenge. To address this, some studies \citep{KumoRFM, GraphFM, OpenGraph} incorporate synthetic data --- either together with real-world data or as an alternative. This includes the use of simple random graph models like stochastic block models \citep{GraphFM}, as well as synthetic graphs generated by large language models \citep{OpenGraph}. However, the majority of graph foundation models rely primarily on real-world datasets for pretraining. The number of datasets used for this purpose varies from as few as one graph \citep{TSGNN}, to larger collections of 2--10 datasets \citep{GCOPE, MDGFM}, and up to several dozens in some studies~\citep{AnyGraph, GraphFM}.

\paragraph{Handling features} One of the key challenges for graph foundation models is handling heterogeneous features that can vary significantly across different datasets. Some approaches address this by focusing exclusively on text-attributed graphs, sometimes additionally converting non-textual features to text, and then applying a text encoder \citep{GFT, UniGraph, OFA}. Methods aiming to deal with arbitrary features often rely on dimensionality reduction techniques such as SVD or PCA to obtain feature embeddings \citep{AnyGraph, GCOPE, MDGFM, SAMGPT}. There are alternatives, such as learning dimension encoding modules that produce feature transformations \citep{FUG}, learning graph patches~\citep{sun2025handling} or replacing node attribute values with their statistical dependencies \citep{STAGE}, but these appear less common. We also highlight \citet{TSGNN}, which constructs separate embeddings for each \texttt{(node, feature)} pair, enabling a more fine-grained representation of feature information.

\paragraph{Handling structure} Handling the structure is more straightforward, as graph neural networks are well-suited for this task, and they are inherently capable of processing arbitrary graph structures. Consequently, many GFMs simply adopt GNNs as their backbone to handle graph structure \citep{GCOPE, TSGNN}. In addition to GNN-based approaches, some methods apply matrix decomposition techniques like SVD to graph-derived matrices (for example, the normalized adjacency matrix or the sum of its powers), to encode structural information \citep{OpenGraph}. However, while GNNs can in principle operate on any graph, their performance may still be limited due to varying graph structures. To address this, some works implement additional mechanisms specifically designed to handle structural differences \citep{SAMGPT, MDGFM}.

\subsection{Limitations of existing GFMs}

\paragraph{Focus on text-attributed graphs} Many existing graph foundation models are specifically designed for text-attributed graphs, where nodes or edges have associated textual information~\citep{GFT, UniGraph, OFA}. These models typically leverage large language models or other text encoders to process textual attributes, integrating natural language representations with graph structures. While this approach can be effective for certain domains such as academic networks or knowledge graphs, it limits the applicability of GFMs across a broader range of graphs where such text attributes are not available. For instance, for graphs representing transportation networks, biological networks, or transaction networks (commonly used for fraud detection tasks), which often come with rich numerical and categorical features, the reliance exclusively on textual information restricts the model's usability and effectiveness. As a result, many current GFMs may not generalize well to graphs with non-textual attributes, hindering their adoption across diverse real-world scenarios.

\paragraph{Limited support for regression tasks} Most publicly available GFMs are designed and evaluated on classification tasks, where the goal is to predict categorical labels for nodes, edges, or entire graphs. To date, no publicly available GFMs, aside from TS-GNN~\citep{TSGNN}, support regression tasks, where the output is a continuous value rather than a class label. This is a substantial limitation because many important graph-based applications require regression instead of classification. The lack of support for regression tasks reduces the practical applicability of current GFMs and highlights an important area for future research.

\paragraph{Misleading use of the ``zero-shot'' term} Some recent studies on graph foundation models have described their methods as operating in a ``zero-shot'' setting~\citep{AnyGraph, OpenGraph}. Typically, these approaches introduce virtual nodes that represent target classes and connect them to the corresponding real nodes with known class labels. Then, the node classification problem reduces to predicting links between the test nodes and the appropriate virtual nodes. This process makes it possible to perform evaluation on unseen graphs without additional finetuning. While inventive and interesting, this technique does not truly realize zero-shot learning. Strictly speaking, zero-shot learning means that no labeled examples of the target classes are available during evaluation. However, the described method requires labeled nodes to be connected to virtual class nodes for effective link prediction. Therefore, the correct term for this setup should be ``in-context learning'', since evaluation does not involve further finetuning but still depends on access to labeled training samples. This inconsistency in terminology may lead to misleading comparisons with baseline approaches. For instance, the aforementioned studies~\citep{AnyGraph, OpenGraph} compare their ``zero-shot'' performance against the one-shot and five-shot results of other baselines, yet they do not clearly report the number of training samples used in ``zero-shot'' evaluation of the proposed method, which makes the comparison harder to interpret.

\subsection{Tabular foundation models} The field of tabular foundation models (TFMs) was pioneered by the TabPFN model \citep{TabPFN}, which was designed to address any tabular problem off-the-shelf. TabPFN employs a transformer-like architecture and works in the in-context learning regime, with the entire downstream training set serving as the prompt. The pretraining of TabPFN was performed on a large number of synthetic datasets designed to mimic typical tabular tasks. The more recent model, TabPFNv2 \citep{TabPFNv2}, employs a more powerful backbone architecture, pretraining on a broader spectrum of synthetic datasets, and advanced techniques of data preprocessing. Nowadays, new TFMs are emerging regularly \citep{ma2024tabdpt, mueller2025mothernet, mitra, LimiX, qu2025tabicl, qu2026tabiclv2, TabPFNv2_5, TabPFNv3}, and their success is exploited beyond the domain of pure tabular tasks, e.g., for time series forecasting \citep{hoo2025tables}. In our work, we demonstrate that TFMs can also serve as a core building block for graph foundation models.

\section{G2T-FM framework}\label{sec:g2t-fm}

\begin{figure*}
\centering
\includegraphics[width=\linewidth]{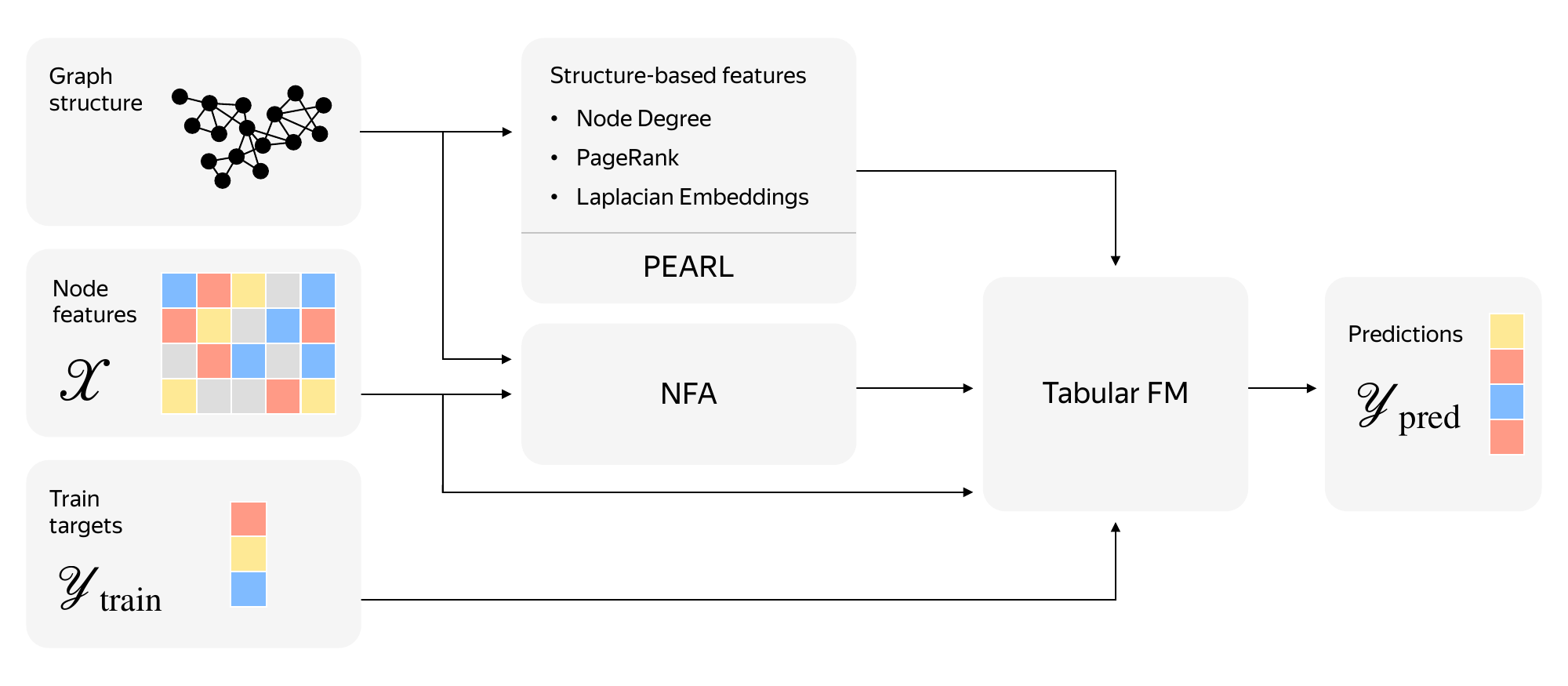}
\caption{Overview of the proposed G2T-FM framework.}
\label{fig:method}
\vspace{-10pt}
\end{figure*}

As discussed above, one of the key challenges for graph foundation models is processing node features that can vary widely across different graphs and domains. To address this, previous approaches primarily rely on one of the following strategies. The first is to apply dimensionality reduction techniques, such as principal component analysis or singular value decomposition~\citep{AnyGraph, GCOPE, MDGFM, SAMGPT}. However, this approach can result in the loss of information, and, although it ensures that all feature vectors share the same dimension after reduction, it remains unclear whether these reduced features are transferable across different graphs. The second strategy employs text encoders to process node or edge features~\citep{GFT, UniGraph, OFA}. This method is highly effective for text-attributed graphs, where the features are naturally represented as text. Nevertheless, many real-world graphs do not include solely textual features~\citep{ivanov2021boost, chen2022does, RelBench, 4DBInfer, GraphLand}. Using text encoders for non-text features can therefore be highly suboptimal, as it does not leverage the nature or structure of these features. Overall, neither of these strategies fully addresses the problem of handling diverse and heterogeneous features. 

Similarly, the problem of adapting to different target spaces has also not been fully solved. Existing approaches, such as converting node classification to link prediction or using textual descriptions of node classes, only work with node classification tasks and not with node regression tasks, which frequently appear in real-world applications of graph machine learning.

However, these challenges are not unique to graph machine learning, and they also arise in tabular machine learning. Recent advances in tabular deep learning, such as the development of foundation models like TabPFNv2~\citep{TabPFNv2}, offer promising solutions for handling diverse feature and target spaces.

We argue that tabular foundation models can be used to create strong models for graph node-level tasks, and, in particular, they can help handle different feature and target spaces. As a proof of concept, we introduce the \underline{G}raph-to-\underline{T}able \underline{F}oundation \underline{M}odel (G2T-FM) framework, which addresses the challenge of learning on graphs with diverse and heterogeneous node features and different targets. Figure~\ref{fig:method} provides an overview of our method.

To process heterogeneous node features, we employ a tabular foundation model like TabPFNv2~\citep{TabPFNv2}, TabPFN-2.5~\citep{TabPFNv2_5} or LimiX~\citep{LimiX}. TFMs are designed for tabular data only, so to make them applicable to graph-structured data, we introduce a graph-based preprocessing step that encodes graph information into the node features. Our goal is to capture the information about different aspects of the graph structure as well as the interplay between the graph structure and the node features. Hence, the new augmented feature vector consists of the original node features and the following graph-based components.

\paragraph{Neighborhood feature aggregation (NFA)} Following \citet{GraphLand}, for each node, we compute aggregated feature statistics over its one-hop neighbors. Namely, for each numerical feature, we compute its mean, maximum, and minimum values over the node's neighbors. For each categorical feature, we first apply one-hot encoding and then compute the mean of the obtained binary features. The computation of NFA uses both the graph structure and node features. This component provides information about the features in the local neighborhood of the node, which is a valuable signal for many graph-related tasks. 

\paragraph{Classic structure-based features (SF)} We also include basic node structural characteristics~--- node degree, PageRank score, and Laplacian eigenvectors. The first two are classic node centrality measures that indicate how ``important'' a particular node is; however, they do so in different ways. Node degree captures strictly local information, while PageRank also captures global information. Then, we compute the first $K$ non-trivial eigenvectors of the symmetric normalized Laplacian and consider the corresponding $K$-dimensional embeddings as additional feature vectors. Laplacian embeddings encode a node's position within the graph relative to other nodes. Thus, such node representations provide valuable information that supplements the centrality measures.

\paragraph{Learnable structure-based encodings (PEARL)} These encodings have been proposed by~\citet{PEARL}. The idea of PEARL is to generate a random value for each node and then apply a GNN using these values as node features. Such random initialization increases the expressive power of GNNs by breaking the structural symmetries. This procedure is repeated $M$ times (each with new random node features) and the resulting node embeddings are averaged so that node permutation equivariance still holds in the limit. \citet{PEARL} propose using this as a learnable module, so that the encodings are trained to improve downstream performance. However, we noticed that PEARL encodings are also useful without training, i.e., when we use a randomly initialized GNN to obtain node embeddings. Thus, this module can produce both non-learnable and learnable representations, where non-learnable representations do not involve parameter optimization (training).

These new features allow us to provide the tabular foundation model with information about many properties of the graph underlying the given dataset. We concatenate them with the original node attributes and use the resulting augmented feature vector as an input to a TFM. The resulting model can be applied in a fully in-context learning regime and, as we show below, it outperforms well-tuned GNNs trained from scratch. In the finetuning regime, we jointly tune PEARL and TFM.

\section{Experimental setup}\label{sec:experimental-setup}

\subsection{Datasets}

In our experiments, we use TabPFNv2 as one of the backbones, which imposes specific constraints that guide our dataset selection. In particular, TabPFNv2 is suitable for both regression tasks and multi-class classification tasks, but the latter are restricted to at most 10 classes. Additionally, it is designed for small-to-medium-scale datasets, requiring no more than roughly 10,000 training samples. As such, we select datasets that fit within these boundaries.

We note that further advances in TFMs have alleviated the 10,000-training-sample limit~\citep{TabPFNv2_5}, and error-correcting output codes~\citep{dietterich1994solving} make it possible to handle more than 10 classes even with TabPFNv2. However, we keep the dataset selection within these boundaries in order to evaluate a broader range of models in their default setups.

To comprehensively assess the capabilities of our method, we construct two collections of datasets. The first focuses on graphs with non-textual node features (our primary setting), while the second contains well-known graph benchmarks with text-based node features. Across these datasets, our selection covers both regression and classification tasks, and includes graphs with both homophilous and non-homophilous structure,\footnote{A graph is called homophilous if its edges tend to connect nodes with similar labels, see \citet{newman2003mixing, platonov2023characterizing, mironov2024revisiting} for details.} as summarized in Table~\ref{tab:dataset-statistics}.

\paragraph{Datasets with non-textual features} 
This collection comprises eight datasets from the GraphLand benchmark~\citep{GraphLand}, all featuring diverse tabular node features.\footnote{By tabular features, we mean a mixture of numerical and categorical features with different distributions.} The datasets include: social networks \texttt{artnet-exp}, \texttt{artnet-views}, and \texttt{twitch-views}; a network of workers from a crowdsourcing platform \texttt{tolokers-2}; a network of users of a review service \texttt{city-reviews}; a co-purchasing network \texttt{hm-prices}; a network of devices \texttt{avazu-ctr}; and a road network \texttt{city-roads-M}.
Importantly, compared to many classical graph datasets, GraphLand provides more diverse and realistic tasks, and thus we mostly focus on this collection in our paper.

\paragraph{Datasets with text-based features} 
This collection includes five classical graph datasets where node features are derived from textual descriptions: a network of users of a question-answering website \texttt{questions}~\citep{platonov2023critical}; a citation network \texttt{pubmed}~\citep{yang2016revisiting}; a social network \texttt{facebook}~\citep{rozemberczki2020characteristic}; a co-purchasing network \texttt{amazon-ratings}~\citep{platonov2023critical}; and a network of Wikipedia pages \texttt{wiki-cs}~\citep{mernyei2020wiki}. 

While specialized graph foundation models that directly process raw text may yield better results for text-attributed graphs, including these datasets allows us to test the generalization of our approach. Also, we explicitly exclude datasets with bag-of-words (BoW) node features, as BoW is less relevant to modern text processing, while introducing additional challenges like high-dimensionality and sparsity.

\begin{table*}[t!]
\caption{The key statistics of the considered graph datasets.}
\label{tab:dataset-statistics}
\begin{center}
\resizebox{0.9\textwidth}{!}{
\begin{tabular}{lrrrrcccc}
\toprule
name & \multicolumn{1}{c}{\# nodes} & \multicolumn{1}{c}{\# edges} & \multicolumn{1}{c}{\# features} & mean degree & \# classes & homophily & feature type \\
\midrule
\texttt{tolokers-2} & $11{,}758$ & $519{,}000$ & $16$ & $88.3$ & 2 & no & tabular \\
\texttt{city-reviews} & $148{,}801$ & $1{,}165{,}415$ & $37$ & $15.7$ & 2 & yes & tabular \\
\texttt{artnet-exp} & $50{,}405$ & $280{,}348$ & $75$ & $11.1$ & 2 & no & tabular \\
\texttt{hm-prices} & $46{,}563$ & $10{,}730{,}995$ & $41$ & $460.9$ & -- & no & tabular \\
\texttt{avazu-ctr} & $76{,}269$ & $10{,}984{,}077$ & $260$ & $288.0$ & -- & no & tabular \\
\texttt{city-roads-M} & $57{,}073$ & $107{,}104$ & $26$ & $3.8$ & -- & yes & tabular \\
\texttt{twitch-views} & $168{,}114$ & $6{,}797{,}557$ & $4$ & $80.9$ & -- & no & tabular \\
\texttt{artnet-views} & $50{,}405$ & $280{,}348$ & $50$ & $11.1$ & -- & no & tabular \\
\texttt{pubmed} & $19{,}717$ & $44{,}324$ & $500$ & $4.5$ & $3$ & yes & text-based \\
\texttt{facebook} & $22{,}470$ & $170{,}823$ & $128$ & $15.2$ & $4$ & yes & text-based \\
\texttt{amazon-ratings} & $24{,}492$ & $93{,}050$ & $300$ & $7.6$ & $5$ & no & text-based \\
\texttt{questions} & $48{,}921$ & $153{,}540$ & $301$ & $6.3$ & 2 & no & text-based \\
\texttt{wiki-cs} & $11{,}701$ & $215{,}603$ & $300$ & $36.9$ & 10 & yes & text-based \\
\bottomrule
\end{tabular}
}
\end{center}
\vspace{-15pt}
\end{table*}

For all datasets, we employ a standardized data splitting protocol, allocating 10\% of the nodes to training, 10\% to validation, and the remaining 80\% to testing. For the GraphLand datasets, we use the official \texttt{RL} (random low) splits. For the remaining datasets, we employ the random stratified splitting, which ensures consistent class distributions across the splits. All the experiments are run in a transductive setting, which is standard for node property prediction in the graph domain. For binary classification tasks, we report average precision. For multiclass classification tasks, we report accuracy. For regression tasks, we report $R^2$. For all metrics, higher is better. 

\subsection{Methods}

In our experiments with G2T-FM, we adopt TabPFNv2~\citep{TabPFNv2}, TabPFN-2.5~\citep{TabPFNv2_5} and LimiX~\citep{LimiX} as backbone models. We refer to these models as G2T-TabPFNv2, G2T-TabPFNv2.5, and G2T-LimiX, respectively. These models are evaluated either in an in-context learning (ICL) or finetuning (FT) settings. 
Following a common practice in tabular PFNs~\citep{TabPFNv2, qu2026tabiclv2}, we use inference-time ensembling for G2T-FM via averaging predictions from 10 forward passes with different random seeds, see Appendix~\ref{appendix:implementation} for details.
For comparison, we also evaluate the following baseline methods. 

First, we include traditional supervised baselines trained from scratch for each dataset. These include four classic GNNs: GCN~\citep{kipf2017semi}, GraphSAGE~\citep{hamilton2017inductive}, GAT~\citep{velivckovic2018graph}, and neighborhood-attention Graph Transformer (GT)~\citep{shi2021masked}.\footnote{Neighborhood-attention GT uses only local attention to node's neighbors, in contrast to graph transformers with global all-to-all attention.} For all GNNs, we use the modifications from \citet{platonov2023critical} that augment GNNs with residual connections~\citep{he2016deep}, layer normalization~\citep{ba2016layer}, and MLP blocks, which often significantly improve their performance. For the GraphLand datasets with tabular features, we also use LightGBM-NFA as a strong baseline~--- which is a popular implementation of gradient-boosted decision trees~\citep{ke2017lightgbm} with input features augmented with graph neighborhood information via NFA~\citep{GraphLand}. The implementation of these models closely resembles that from GraphLand~\citep{GraphLand} in terms of both model architecture and hyperparameter tuning, see Appendix \ref{appendix:implementation} for more details.

Second, we employ several publicly available graph foundation models. Despite significant interest in developing GFMs recently, most of the research is focused exclusively on text-attributed graphs (as discussed above), so we were able to find only a few openly available models that support node property prediction in graphs with arbitrary node feature spaces. Specifically, we employ AnyGraph~\citep{AnyGraph}, OpenGraph~\citep{OpenGraph}, and TS-GNN~\citep{TSGNN}, which are used in the in-context learning (ICL) regime, as well as GCOPE~\citep{GCOPE}, SAMGPT~\citep{SAMGPT} and MDGFM~\citep{MDGFM}, which are used in the finetuning (FT) regime. For all these methods, we use the original implementations provided by the authors. Further, we were only able to evaluate these methods on datasets with node classification tasks, as none of them support node regression.\footnote{While TS-GNN can in theory perform node regression, its current official implementation does not support that.}

For all the methods, we run each experiment 10 times (5 times for GFMs from prior literature) and report the mean and standard deviation of the model performance, since the results are affected by stochasticity during model training and inference: some of the considered baselines have stochastic inference by design, while others require training or finetuning with different random states.

\section{Experimental results}

\begin{table*}[t!]
\caption{Evaluation results on datasets with tabular features, i.e., datasets from the GraphLand benchmark under the \texttt{RL} (random low) data split. For each column, we highlight \textcolor{red}{first}, \textcolor{blue}{second}, and \textcolor{violet}{third} best results with a color.
We also report average rank over classification datasets in the `AR (cls)' column and over all datasets in the `AR (all)' column.
During the computation of `AR (all)' we exclude all methods that do not support regression tasks, so that ranks are computed only over tables with no missing entries.
}
\label{tab:graphland}
\begin{center}
\resizebox{\textwidth}{!}{
\begin{tabular}{lcccccccccc}
\toprule
 & \texttt{artnet-exp} & \texttt{city-reviews} & \texttt{tolokers-2} & \texttt{artnet-views} & \texttt{avazu-ctr} & \texttt{city-roads-M} & \texttt{hm-prices} & \texttt{twitch-views} & \textbf{AR (cls)} & \textbf{AR (all)} \\
\midrule
LightGBM-NFA & $46.13 \pm 0.04$ & $78.53 \pm 0.01$ & $56.34 \pm 0.06$ & $56.10 \pm 0.02$ & $31.71 \pm 0.01$ & $61.18 \pm 0.03$ & $70.84 \pm 0.04$ & $60.14 \pm 0.01$ & $\mathbf{6.33}$ & $\mathbf{6.62}$ \\
GCN & $44.86 \pm 0.36$ & $77.81 \pm 0.15$ & $56.27 \pm 0.31$ & $56.03 \pm 0.25$ & $32.00 \pm 0.16$ & $58.82 \pm 0.25$ & $68.02 \pm 0.42$ & \textcolor{red}{$75.51 \pm 0.05$} & $\mathbf{8.67}$ & $\mathbf{7.38}$ \\
GraphSAGE & $45.14 \pm 0.36$ & $78.17 \pm 0.10$ & $54.43 \pm 0.34$ & $49.32 \pm 0.91$ & $31.44 \pm 0.16$ & $59.44 \pm 0.27$ & $70.00 \pm 0.74$ & $66.29 \pm 0.32$ & $\mathbf{8.00}$ & $\mathbf{8.38}$ \\
GAT & $45.06 \pm 0.52$ & $77.74 \pm 0.21$ & $57.41 \pm 0.85$ & $53.60 \pm 0.24$ & $32.63 \pm 0.17$ & $59.86 \pm 0.20$ & $72.07 \pm 1.22$ & $72.89 \pm 0.27$ & $\mathbf{7.67}$ & $\mathbf{6.50}$ \\
GT & $46.41 \pm 0.71$ & $77.34 \pm 0.21$ & $56.98 \pm 0.55$ & $53.37 \pm 0.46$ & $31.11 \pm 0.49$ & $59.55 \pm 0.28$ & $69.44 \pm 0.94$ & $72.13 \pm 0.13$ & $\mathbf{7.00}$ & $\mathbf{7.50}$ \\
\midrule
OpenGraph (ICL) & $15.16 \pm 0.83$ & $59.09 \pm 0.72$ & $40.38 \pm 1.13$ & -- & -- & -- & -- & -- & $\mathbf{12.67}$ & -- \\
AnyGraph (ICL) & $12.84 \pm 0.93$ & $63.71 \pm 1.45$ & $28.75 \pm 3.56$ & -- & -- & -- & -- & -- & $\mathbf{14.67}$ & -- \\
TS-GNN (ICL) & $20.44 \pm 1.05$ & $43.46 \pm 5.17$ & $38.54 \pm 0.94$ & -- & -- & -- & -- & -- & $\mathbf{12.67}$ & -- \\
GCOPE (FT) & $14.92 \pm 1.56$ & $67.16 \pm 0.98$ & $28.81 \pm 1.28$ & -- & -- & -- & -- & -- & $\mathbf{13.67}$ & -- \\
SAMGPT (FT) & $16.20 \pm 0.93$ & $45.91 \pm 2.68$ & $36.93 \pm 1.07$ & -- & -- & -- & -- & -- & $\mathbf{13.33}$ & -- \\
MDGFM (FT) & $16.99 \pm 1.75$ & $31.24 \pm 8.50$ & $31.97 \pm 1.34$ & -- & -- & -- & -- & -- & $\mathbf{14.00}$ & -- \\
\midrule
G2T-TabPFNv2 (ICL) & $45.73 \pm 0.03$ & $77.28 \pm 0.25$ & \textcolor{violet}{$60.39 \pm 0.19$} & $59.72 \pm 0.14$ & $28.00 \pm 0.36$ & $60.12 \pm 0.03$ & $65.75 \pm 0.02$ & $70.21 \pm 0.04$ & $\mathbf{6.67}$ & $\mathbf{7.38}$ \\
G2T-TabPFNv2.5 (ICL) & \textcolor{blue}{$49.33 \pm 0.12$} & \textcolor{blue}{$79.99 \pm 0.04$} & \textcolor{blue}{$61.21 \pm 0.12$} & \textcolor{violet}{$60.64 \pm 0.09$} & \textcolor{violet}{$33.06 \pm 0.11$} & \textcolor{violet}{$63.53 \pm 0.07$} & \textcolor{violet}{$74.32 \pm 0.09$} & $71.64 \pm 0.06$ & \textcolor{red}{$\mathbf{2.00}$} & \textcolor{blue}{$\mathbf{3.00}$} \\
G2T-LimiX (ICL) & \textcolor{violet}{$48.42 \pm 0.78$} & \textcolor{violet}{$78.98 \pm 0.44$} & \textcolor{red}{$61.60 \pm 0.18$} & \textcolor{blue}{$61.58 \pm 0.08$} & $32.70 \pm 0.14$ & \textcolor{blue}{$65.16 \pm 0.07$} & \textcolor{blue}{$76.14 \pm 0.08$} & $71.31 \pm 0.06$ & \textcolor{violet}{$\mathbf{2.33}$} & \textcolor{blue}{$\mathbf{3.00}$} \\
\midrule
G2T-TabPFNv2 (FT) & $47.53 \pm 0.62$ & $78.69 \pm 0.30$ & $57.54 \pm 0.93$ & $60.60 \pm 0.16$ & \textcolor{blue}{$33.38 \pm 0.41$} & $63.05 \pm 0.39$ & $72.58 \pm 0.40$ & \textcolor{violet}{$74.24 \pm 0.11$} & $\mathbf{4.33}$ & $\mathbf{3.75}$ \\
G2T-LimiX (FT) & \textcolor{red}{$50.39 \pm 0.19$} & \textcolor{red}{$80.65 \pm 0.05$} & $59.75 \pm 1.14$ & \textcolor{red}{$63.24 \pm 0.07$} & \textcolor{red}{$34.09 \pm 0.37$} & \textcolor{red}{$66.29 \pm 0.12$} & \textcolor{red}{$77.37 \pm 0.17$} & \textcolor{blue}{$74.91 \pm 0.06$} & \textcolor{red}{$\mathbf{2.00}$} & \textcolor{red}{$\mathbf{1.50}$} \\
\bottomrule
\end{tabular}
}
\end{center}
\vspace{-10pt}
\end{table*}

Table~\ref{tab:graphland} contains the evaluation results on graph datasets with tabular features, and Table~\ref{tab:classic} contains the additional results on datasets with text-derived features. In addition to the results on individual datasets, we also report the average ranks (AR). Below, we summarize and discuss our key observations.

\begin{observation}
In our evaluation, the existing publicly available graph foundation models perform substantially worse than well-tuned traditional GNNs trained from scratch.
\end{observation}

This observation holds true across both collections of datasets we evaluated. The sole exception we identified is the performance of TS-GNN on the \texttt{amazon-ratings} dataset, where it surpassed the GNNs trained from scratch. In all other cases, the performance of the existing GFMs was significantly lower than that of our GNN baselines.

While many GFM publications report outperforming traditional GNNs, our findings suggest otherwise. We hypothesize that this discrepancy stems from several key differences in the evaluation protocol. First, unlike some GFM studies that focus on few-shot benchmarks, we use larger, more realistic, training splits (10\%/10\%/80\%) that allow GNNs to be trained effectively from scratch. Second, we ensure our GNN baselines are highly competitive by performing a thorough hyperparameter optimization and using the GNN architectures from~\citet{platonov2023critical} that include established performance-enhancing features like residual connections and normalization~\citep{CGASB, CGASB2}, which are often absent in simpler baselines.

\begin{table*}[t!]
\caption{Evaluation results on classic graph datasets with text-based features. For each column, we highlight \textcolor{red}{first}, \textcolor{blue}{second}, and \textcolor{violet}{third} best results with a color.
We also report average rank (AR) in the last column.
}
\label{tab:classic}
\begin{center}
\resizebox{0.8\textwidth}{!}{
\begin{tabular}{lcccccc}
\toprule
 & \texttt{amazon-ratings} & \texttt{facebook} & \texttt{pubmed} & \texttt{questions} & \texttt{wiki-cs} & \textbf{AR} \\
\midrule
GCN & $41.43 \pm 0.49$ & $91.26 \pm 0.21$ & $85.46 \pm 0.19$ & $15.42 \pm 0.67$ & $81.74 \pm 0.21$ & $\mathbf{7.00}$ \\
GraphSAGE & $40.07 \pm 0.53$ & $91.12 \pm 0.22$ & $86.04 \pm 0.27$ & $16.55 \pm 0.64$ & $81.50 \pm 0.27$ & $\mathbf{7.60}$ \\
GAT & $40.67 \pm 0.55$ & \textcolor{blue}{$92.61 \pm 0.21$} & $84.81 \pm 0.24$ & $16.75 \pm 0.67$ & \textcolor{violet}{$82.25 \pm 0.27$} & $\mathbf{5.80}$ \\
GT & $41.56 \pm 0.40$ & $91.71 \pm 0.22$ & $84.95 \pm 0.19$ & $14.03 \pm 0.90$ & \textcolor{blue}{$82.54 \pm 0.21$} & $\mathbf{6.20}$ \\
\midrule
OpenGraph (ICL) & $29.36 \pm 1.24$ & $75.27 \pm 5.05$ & $70.30 \pm 2.67$ & $3.77 \pm 0.65$ & $75.66 \pm 0.39$ & $\mathbf{12.60}$ \\
AnyGraph (ICL) & $33.49 \pm 3.44$ & $61.17 \pm 8.64$ & $65.31 \pm 6.26$ & $4.27 \pm 0.66$ & $65.17 \pm 2.51$ & $\mathbf{13.20}$ \\
TS-GNN (ICL) & $43.00 \pm 0.13$ & $77.87 \pm 2.73$ & $64.41 \pm 5.11$ & $5.00 \pm 0.48$ & $46.25 \pm 9.77$ & $\mathbf{12.00}$ \\
GCOPE (FT) & $39.90 \pm 0.43$ & $85.08 \pm 0.17$ & $79.35 \pm 0.70$ & $6.59 \pm 0.43$ & $59.13 \pm 1.20$ & $\mathbf{11.20}$ \\
SAMGPT (FT) & $21.20 \pm 1.90$ & $80.61 \pm 0.93$ & $71.93 \pm 1.45$ & $8.24 \pm 0.48$ & $63.82 \pm 2.35$ & $\mathbf{12.00}$ \\
MDGFM (FT) & $21.37 \pm 1.34$ & $70.30 \pm 3.59$ & $67.54 \pm 1.94$ & $10.55 \pm 0.50$ & $58.71 \pm 3.74$ & $\mathbf{13.00}$ \\
\midrule
G2T-TabPFNv2 (ICL) & $41.99 \pm 0.12$ & $91.27 \pm 0.12$ & \textcolor{violet}{$89.39 \pm 0.15$} & $18.46 \pm 0.58$ & $80.09 \pm 0.26$ & $\mathbf{6.00}$ \\
G2T-TabPFNv2.5 (ICL) & $43.90 \pm 0.17$ & $91.56 \pm 0.08$ & $89.06 \pm 0.23$ & \textcolor{blue}{$21.31 \pm 0.12$} & $81.68 \pm 0.13$ & \textcolor{violet}{$\mathbf{4.40}$} \\
G2T-LimiX (ICL) & \textcolor{violet}{$45.03 \pm 0.10$} & $91.72 \pm 0.10$ & $89.25 \pm 0.14$ & \textcolor{violet}{$20.97 \pm 0.52$} & $80.93 \pm 0.13$ & \textcolor{violet}{$\mathbf{4.40}$} \\
\midrule
G2T-TabPFNv2 (FT) & \textcolor{blue}{$45.19 \pm 0.25$} & \textcolor{violet}{$92.53 \pm 0.05$} & \textcolor{blue}{$90.52 \pm 0.21$} & $18.91 \pm 1.06$ & $81.46 \pm 0.29$ & \textcolor{blue}{$\mathbf{3.60}$} \\
G2T-LimiX (FT) & \textcolor{red}{$46.09 \pm 0.23$} & \textcolor{red}{$92.83 \pm 0.12$} & \textcolor{red}{$90.94 \pm 0.09$} & \textcolor{red}{$22.98 \pm 0.27$} & \textcolor{red}{$83.38 \pm 0.17$} & \textcolor{red}{$\mathbf{1.00}$} \\
\bottomrule
\end{tabular}
}
\end{center}
\vspace{-10pt}
\end{table*}

\begin{observation}
When endowed with a strong TFM backbone, G2T-FM evaluated in the in-context learning mode performs better on average than traditional GNNs trained from scratch.
\end{observation}

In particular, in terms of the average rank on both collections of datasets, G2T-LimiX and G2T-TabPFNv2.5 outperform all baselines trained from scratch, while G2T-TabPFNv2 performs on par with them. Results of G2T-FM on individual datasets are also strong. For example, on \texttt{artnet-exp}, \texttt{tolokers-2}, \texttt{artnet-views}, \texttt{city-roads-M}, and \texttt{hm-prices} G2T-LimiX surpasses all traditional baselines with more than two percentage points margin.

\begin{observation}
After finetuning, G2T-FM outperforms on average all traditional baselines trained from scratch on both collections of datasets with G2T-LimiX achieving especially strong improvements.
\end{observation}

On datasets with tabular features, G2T-LimiX achieves not only the best average rank, but also the best results on six out of eight datasets, often yielding notable improvements of more than two percentage points compared to the best GNN result.

On datasets with text-based features, finetuned G2T-LimiX is also strong, achieving the best results on all the considered datasets, often with noticeable improvements over traditional GNNs. We note, however, that specialized models that directly process raw text may achieve higher accuracy on text-attributed graphs. Though, comparing against such models is outside the scope of this work.

\section{Ablations}\label{sec:ablation}

\paragraph{G2T-FM components} First, we provide an ablation of the G2T-FM components by removing them from G2T-FM and comparing performance. Table~\ref{tab:g2t_fm-ablation} shows the results of this ablation, from which we conclude that all the components are critical for the performance of G2T-FM. In particular, neighborhood feature aggregation (NFA) and classic structure-based features (SF) improve the overall performance of G2T-FM, while PEARL allows one to drastically improve performance in rare cases where standard augmented features are not sufficient.

\paragraph{Augmenting baselines with the same components} Second, one may argue that the performance improvements of G2T-FM come solely from the fact that it employs augmented features that are not accessible to the GNN and LightGBM baselines. To verify this, we provide the baselines with exactly the same features as G2T-FM. The results are presented in Table~\ref{tab:extended-eval}.
While some improvements achieved by G2T-FM can be explained by its access to the features that are not used by traditional GNNs, G2T-FM shows strong performance even against the enhanced baselines. In particular, G2T-LimiX achieves better average rank, and on some datasets it outperforms all other methods by a notable margin.

\paragraph{Evaluation without ensembling} The default evaluation setup of G2T-FM uses an ensemble over 10 forward passes.
For completeness, Table~\ref{tab:extended-eval} reports the results obtained without ensembling. 
Overall, ensembling yields small but consistent gains, with larger improvements in certain cases, e.g., G2T-TabPFNv2.5 (ICL) on \texttt{artnet-exp} and \texttt{hm-prices} or G2T-LimiX (FT) on \texttt{artnet-views}. Nevertheless, G2T-FM remains competitive without ensembling, still outperforming traditional supervised baselines on average even in an in-context learning regime.

\paragraph{Summary} The gains of G2T-FM come from the synergy between the TFM backbone and our graph-to-table components. The ablations show that removing any component degrades the performance on average, and providing the baselines with the same augmented features does not close the gap. Hence, both the backbone and the proposed components are necessary for the strong performance.

\begin{table*}[t!]
\caption{Ablation of the components of G2T-FM. `SF' stands for Structure-based Features, which include degree, PageRank, and Laplacian eigenvectors.}
\label{tab:g2t_fm-ablation}
\vspace{-4pt}
\begin{center}
\resizebox{\textwidth}{!}{
\begin{tabular}{lccccccccc}
\toprule
 & \texttt{artnet-exp} & \texttt{city-reviews} & \texttt{tolokers-2} & \texttt{artnet-views} & \texttt{avazu-ctr} & \texttt{city-roads-M} & \texttt{hm-prices} & \texttt{twitch-views} & \textbf{AR} \\
\midrule
LightGBM-NFA & $46.13 \pm 0.04$ & $78.53 \pm 0.01$ & $56.34 \pm 0.06$ & $56.10 \pm 0.02$ & $31.71 \pm 0.01$ & $61.18 \pm 0.03$ & $70.84 \pm 0.04$ & $60.14 \pm 0.01$ & $\mathbf{12.25}$ \\
GCN & $44.86 \pm 0.36$ & $77.81 \pm 0.15$ & $56.27 \pm 0.31$ & $56.03 \pm 0.25$ & $32.00 \pm 0.16$ & $58.82 \pm 0.25$ & $68.02 \pm 0.42$ & $75.51 \pm 0.05$ & $\mathbf{12.12}$ \\
GraphSAGE & $45.14 \pm 0.36$ & $78.17 \pm 0.10$ & $54.43 \pm 0.34$ & $49.32 \pm 0.91$ & $31.44 \pm 0.16$ & $59.44 \pm 0.27$ & $70.00 \pm 0.74$ & $66.29 \pm 0.32$ & $\mathbf{13.50}$ \\
GAT & $45.06 \pm 0.52$ & $77.74 \pm 0.21$ & $57.41 \pm 0.85$ & $53.60 \pm 0.24$ & $32.63 \pm 0.17$ & $59.86 \pm 0.20$ & $72.07 \pm 1.22$ & $72.89 \pm 0.27$ & $\mathbf{11.00}$ \\
GT & $46.41 \pm 0.71$ & $77.34 \pm 0.21$ & $56.98 \pm 0.55$ & $53.37 \pm 0.46$ & $31.11 \pm 0.49$ & $59.55 \pm 0.28$ & $69.44 \pm 0.94$ & $72.13 \pm 0.13$ & $\mathbf{12.50}$ \\
\midrule
G2T-LimiX (ICL) & $48.42 \pm 0.78$ & $78.98 \pm 0.44$ & $61.60 \pm 0.18$ & $61.58 \pm 0.08$ & $32.70 \pm 0.14$ & $65.16 \pm 0.07$ & $76.14 \pm 0.08$ & $71.31 \pm 0.06$ & $\mathbf{6.12}$ \\
\midrule
w/o NFA (ICL) & $47.17 \pm 0.11$ & $79.18 \pm 0.14$ & $61.30 \pm 0.20$ & $61.24 \pm 0.07$ & $32.14 \pm 0.11$ & $63.11 \pm 0.10$ & $71.27 \pm 0.18$ & $70.63 \pm 0.06$ & $\mathbf{8.38}$ \\
w/o SF \& PEARL (ICL) & $48.33 \pm 0.50$ & $79.24 \pm 0.28$ & $58.57 \pm 0.14$ & $58.44 \pm 0.06$ & $32.87 \pm 0.08$ & $65.64 \pm 0.08$ & $76.24 \pm 0.11$ & $60.77 \pm 0.04$ & $\mathbf{7.88}$ \\
w/o SF (ICL) & $48.26 \pm 0.60$ & $79.24 \pm 0.20$ & $58.29 \pm 0.12$ & $59.22 \pm 0.05$ & $33.14 \pm 0.07$ & $65.99 \pm 0.07$ & $76.03 \pm 0.11$ & $68.64 \pm 0.02$ & $\mathbf{7.38}$ \\
w/o PEARL (ICL) & $48.40 \pm 0.77$ & $78.87 \pm 0.52$ & $61.72 \pm 0.21$ & $61.34 \pm 0.09$ & $32.58 \pm 0.15$ & $64.85 \pm 0.07$ & $76.33 \pm 0.08$ & $66.58 \pm 0.05$ & $\mathbf{7.00}$ \\
\midrule
G2T-LimiX (FT) & $50.39 \pm 0.19$ & $80.65 \pm 0.05$ & $59.75 \pm 1.14$ & $63.24 \pm 0.07$ & $34.09 \pm 0.37$ & $66.29 \pm 0.12$ & $77.37 \pm 0.17$ & $74.91 \pm 0.06$ & $\mathbf{2.75}$ \\
\midrule
w/o NFA (FT) & $47.24 \pm 0.38$ & $80.02 \pm 0.06$ & $60.52 \pm 0.67$ & $62.53 \pm 0.09$ & $33.84 \pm 0.23$ & $64.18 \pm 0.29$ & $73.28 \pm 0.34$ & $74.36 \pm 0.08$ & $\mathbf{6.00}$ \\
w/o SF \& PEARL (FT) & $49.88 \pm 0.16$ & $80.26 \pm 0.10$ & $58.16 \pm 0.65$ & $60.51 \pm 0.08$ & $34.90 \pm 0.23$ & $66.25 \pm 0.09$ & $77.54 \pm 0.26$ & $62.40 \pm 0.06$ & $\mathbf{5.62}$ \\
w/o SF (FT) & $50.01 \pm 0.14$ & $80.48 \pm 0.03$ & $58.05 \pm 0.49$ & $60.84 \pm 0.14$ & $34.23 \pm 0.48$ & $66.38 \pm 0.10$ & $77.29 \pm 0.41$ & $73.24 \pm 0.16$ & $\mathbf{4.38}$ \\
w/o PEARL (FT) & $50.13 \pm 0.14$ & $80.66 \pm 0.07$ & $60.62 \pm 0.90$ & $63.25 \pm 0.07$ & $34.92 \pm 0.17$ & $66.29 \pm 0.13$ & $77.38 \pm 0.32$ & $67.92 \pm 0.05$ & $\mathbf{2.88}$ \\
\bottomrule
\end{tabular}
}
\end{center}
\vspace{-10pt}
\end{table*}

\begin{table*}[t!]
\caption{
Extended evaluation of G2T-FM.
`M' stands for Modified and means that we add NFA, classic structure-based features, and PEARL encodings to their features.
`L' (Light) indicates that G2T-FM was evaluated without ensembling.
}
\label{tab:extended-eval}
\vspace{-4pt}
\begin{center}
\resizebox{\textwidth}{!}{
\begin{tabular}{lccccccccc}
\toprule
 & \texttt{artnet-exp} & \texttt{city-reviews} & \texttt{tolokers-2} & \texttt{artnet-views} & \texttt{avazu-ctr} & \texttt{city-roads-M} & \texttt{hm-prices} & \texttt{twitch-views} & \textbf{AR} \\
\midrule
LightGBM-NFA & $46.13 \pm 0.04$ & $78.53 \pm 0.01$ & $56.34 \pm 0.06$ & $56.10 \pm 0.02$ & $31.71 \pm 0.01$ & $61.18 \pm 0.03$ & $70.84 \pm 0.04$ & $60.14 \pm 0.01$ & $\mathbf{13.12}$ \\
GCN & $44.86 \pm 0.36$ & $77.81 \pm 0.15$ & $56.27 \pm 0.31$ & $56.03 \pm 0.25$ & $32.00 \pm 0.16$ & $58.82 \pm 0.25$ & $68.02 \pm 0.42$ & $75.51 \pm 0.05$ & $\mathbf{14.38}$ \\
GraphSAGE & $45.14 \pm 0.36$ & $78.17 \pm 0.10$ & $54.43 \pm 0.34$ & $49.32 \pm 0.91$ & $31.44 \pm 0.16$ & $59.44 \pm 0.27$ & $70.00 \pm 0.74$ & $66.29 \pm 0.32$ & $\mathbf{16.25}$ \\
GAT & $45.06 \pm 0.52$ & $77.74 \pm 0.21$ & $57.41 \pm 0.85$ & $53.60 \pm 0.24$ & $32.63 \pm 0.17$ & $59.86 \pm 0.20$ & $72.07 \pm 1.22$ & $72.89 \pm 0.27$ & $\mathbf{12.75}$ \\
GT & $46.41 \pm 0.71$ & $77.34 \pm 0.21$ & $56.98 \pm 0.55$ & $53.37 \pm 0.46$ & $31.11 \pm 0.49$ & $59.55 \pm 0.28$ & $69.44 \pm 0.94$ & $72.13 \pm 0.13$ & $\mathbf{15.12}$ \\
\midrule
LightGBM-NFA (M) & $45.57 \pm 0.20$ & $78.68 \pm 0.05$ & $57.16 \pm 0.74$ & $57.53 \pm 0.05$ & $31.31 \pm 0.09$ & $60.86 \pm 0.17$ & $70.25 \pm 0.15$ & $65.17 \pm 0.04$ & $\mathbf{13.38}$ \\
GCN (M) & $43.44 \pm 0.34$ & $77.07 \pm 0.28$ & $58.71 \pm 0.48$ & $56.14 \pm 0.25$ & $31.10 \pm 0.23$ & $57.91 \pm 0.24$ & $70.73 \pm 0.28$ & \textcolor{red}{$77.11 \pm 0.10$} & $\mathbf{14.25}$ \\
GraphSAGE (M) & $44.31 \pm 0.56$ & $77.95 \pm 0.10$ & $59.59 \pm 0.54$ & $55.39 \pm 0.34$ & $31.51 \pm 0.43$ & $59.66 \pm 0.10$ & $70.50 \pm 0.50$ & $75.93 \pm 0.20$ & $\mathbf{12.62}$ \\
GAT (M) & $44.36 \pm 0.53$ & $77.47 \pm 0.15$ & $57.76 \pm 0.74$ & $56.51 \pm 0.37$ & $31.97 \pm 0.24$ & $59.57 \pm 0.45$ & $72.46 \pm 0.51$ & \textcolor{blue}{$77.04 \pm 0.08$} & $\mathbf{11.75}$ \\
GT (M) & $43.03 \pm 0.63$ & $76.43 \pm 0.11$ & $58.79 \pm 0.80$ & $56.39 \pm 0.33$ & $29.86 \pm 0.71$ & $59.85 \pm 0.43$ & $71.84 \pm 0.68$ & \textcolor{violet}{$76.15 \pm 0.12$} & $\mathbf{13.62}$ \\
\midrule
G2T-TabPFNv2 (ICL, L) & $45.69 \pm 0.05$ & $77.19 \pm 0.26$ & $60.25 \pm 0.21$ & $59.69 \pm 0.18$ & $28.06 \pm 0.19$ & $60.09 \pm 0.10$ & $65.74 \pm 0.04$ & $69.91 \pm 0.10$ & $\mathbf{14.38}$ \\
G2T-TabPFNv2.5 (ICL, L) & $47.89 \pm 0.08$ & $79.34 \pm 0.04$ & $60.79 \pm 0.32$ & $60.01 \pm 0.10$ & $32.69 \pm 0.07$ & $62.65 \pm 0.24$ & $72.59 \pm 0.20$ & $70.78 \pm 0.03$ & $\mathbf{7.38}$ \\
G2T-LimiX (ICL, L) & $48.44 \pm 0.23$ & $77.29 \pm 0.54$ & \textcolor{violet}{$61.13 \pm 0.21$} & $60.95 \pm 0.09$ & $32.41 \pm 0.12$ & $64.53 \pm 0.09$ & $75.41 \pm 0.04$ & $71.08 \pm 0.07$ & $\mathbf{7.25}$ \\
\midrule
G2T-TabPFNv2 (ICL) & $45.73 \pm 0.03$ & $77.28 \pm 0.25$ & $60.39 \pm 0.19$ & $59.72 \pm 0.14$ & $28.00 \pm 0.36$ & $60.12 \pm 0.03$ & $65.75 \pm 0.02$ & $70.21 \pm 0.04$ & $\mathbf{13.62}$ \\
G2T-TabPFNv2.5 (ICL) & \textcolor{violet}{$49.33 \pm 0.12$} & \textcolor{violet}{$79.99 \pm 0.04$} & \textcolor{blue}{$61.21 \pm 0.12$} & $60.64 \pm 0.09$ & $33.06 \pm 0.11$ & $63.53 \pm 0.07$ & $74.32 \pm 0.09$ & $71.64 \pm 0.06$ & \textcolor{violet}{$\mathbf{4.88}$} \\
G2T-LimiX (ICL) & $48.42 \pm 0.78$ & $78.98 \pm 0.44$ & \textcolor{red}{$61.60 \pm 0.18$} & \textcolor{violet}{$61.58 \pm 0.08$} & $32.70 \pm 0.14$ & \textcolor{violet}{$65.16 \pm 0.07$} & \textcolor{violet}{$76.14 \pm 0.08$} & $71.31 \pm 0.06$ & \textcolor{violet}{$\mathbf{4.88}$} \\
\midrule
G2T-TabPFNv2 (FT, L) & $46.68 \pm 0.71$ & $78.82 \pm 0.14$ & $57.85 \pm 1.30$ & $60.59 \pm 0.12$ & $32.77 \pm 0.54$ & $63.02 \pm 0.38$ & $72.50 \pm 0.72$ & $74.25 \pm 0.15$ & $\mathbf{7.62}$ \\
G2T-LimiX (FT, L) & \textcolor{blue}{$49.79 \pm 0.21$} & \textcolor{blue}{$80.13 \pm 0.13$} & $60.82 \pm 0.57$ & \textcolor{blue}{$62.08 \pm 0.12$} & \textcolor{blue}{$34.03 \pm 0.35$} & \textcolor{blue}{$65.87 \pm 0.11$} & \textcolor{blue}{$76.72 \pm 0.20$} & $74.31 \pm 0.13$ & \textcolor{blue}{$\mathbf{2.88}$} \\
\midrule
G2T-TabPFNv2 (FT) & $47.53 \pm 0.62$ & $78.69 \pm 0.30$ & $57.54 \pm 0.93$ & $60.60 \pm 0.16$ & \textcolor{violet}{$33.38 \pm 0.41$} & $63.05 \pm 0.39$ & $72.58 \pm 0.40$ & $74.24 \pm 0.11$ & $\mathbf{7.38}$ \\
G2T-LimiX (FT) & \textcolor{red}{$50.39 \pm 0.19$} & \textcolor{red}{$80.65 \pm 0.05$} & $59.75 \pm 1.14$ & \textcolor{red}{$63.24 \pm 0.07$} & \textcolor{red}{$34.09 \pm 0.37$} & \textcolor{red}{$66.29 \pm 0.12$} & \textcolor{red}{$77.37 \pm 0.17$} & $74.91 \pm 0.06$ & \textcolor{red}{$\mathbf{2.50}$} \\
\bottomrule
\end{tabular}
}
\end{center}
\vspace{-10pt}
\end{table*}

\section{Conclusion}

In this work, we have shown that tabular foundation models can be successfully employed for solving graph problems since they are able to process heterogeneous feature and target spaces. To show this, we proposed a simple G2T-FM framework, which converts a graph task into a tabular task by augmenting the node features with graph information, and then applies a tabular foundation model. Our empirical results show the strong performance of G2T-FM, both in the in-context learning and finetuning regimes. In particular, G2T-LimiX and G2T-TabPFNv2.5 outperform all well-tuned GNN baselines in terms of the average rank even in the in-context learning mode. After finetuning, G2T-LimiX achieves the best results on most of the datasets, often outperforming the best GNN by a significant margin. Our approach is simple, but it provides significant advantages over prior attempts at developing GFMs and shows the potential of creating truly generalizable GFMs that can achieve strong results across diverse tasks and real-world applications of graph machine learning, regardless of the inherent feature and target spaces. Importantly, G2T-FM can be combined with any tabular foundation model. Thus, future advancements in TFMs can be easily transferred to the graph domain, providing a strong baseline for future GFM studies. 
Despite its strong performance, the G2T-FM framework is only a first step towards utilizing models and ideas from the tabular domain for developing truly generalizable GFMs. Hence, our work has several noteworthy limitations that suggest future research directions, which we discuss in Appendix~\ref{appendix:limitations}.


\bibliographystyle{refstyle}
\bibliography{references}

\newpage
\appendix

\section{Limitations and future work}\label{appendix:limitations}

Despite its strong performance, the proposed G2T-FM framework is only a first step towards utilizing models and ideas from the tabular domain for developing truly generalizable GFMs. Hence, our work has several noteworthy limitations that suggest future research directions.

First, the present version of G2T-FM uses only basic methods for processing graph structures. Future research can bring more graph-specific components into the framework, such as more complex aggregation mechanisms (including learnable and multi-hop aggregations) and cross-graph pretraining. We believe these extensions could enable the model to better capture graph-specific information and transfer knowledge across different graphs. We also note that our preliminary experiments with multi-hop NFA did not provide consistent improvements.

Second, G2T-FM inherits several restrictions from its TFM backbone. In particular, when endowed with TabPFNv2, G2T-FM cannot handle classification datasets with more than 10 classes, and its training set size is limited to at most 10{,}000 samples. These limitations currently make it difficult to apply G2T-FM to large-scale datasets. However, future research on tabular foundation models can alleviate these limitations and new TFMs can be directly used within the G2T-FM framework.

Finally, the scope of our work is limited to node-level prediction tasks. Potentially, G2T-FM can be applied to link-level tasks, but its performance on such problems requires further analysis. Regarding graph-level tasks, adapting G2T-FM is less straightforward. 

\section{Implementation details}\label{appendix:implementation}

In this section, we describe additional implementation details. Further information and the full code are available in our repository: \url{https://github.com/yandex-research/g2t-fm}.

\subsection{G2T-FM}

\paragraph{Finetuning} For the finetuning experiments, we follow the procedure outlined by \citet{rubachev2025finetuning}. Rather than using parameter-efficient finetuning, we opt for full model finetuning, as previous work indicates this yields better performance. We search for the optimal learning rate over the logarithmic grid of 10 values, ranging from $5 \times 10^{-6}$ to $5 \times 10^{-4}$.

\paragraph{Ensembling} By default, we evaluate G2T-FM using an ensemble of 10 members. Ensemble members share the same model weights, but apply possibly different preprocessing schemes to the input features (e.g., column permutation) and targets (e.g., label shuffling for multiclass classification tasks). We employ ensembling in both in-context learning and finetuning settings.
For each optimization step during finetuning, we sample a single random ensemble member (i.e., one preprocessing scheme) to compute the gradient. This strategy yields the advantages of ensembling while limiting computational costs. Specifically, because intermediate evaluations during finetuning occur only once every 10 gradient steps, evaluating the full ensemble requires just 9 additional forward passes for every 10 forward-backward passes. Therefore, the maximum time overhead of our ensembling approach with this strategy is at most double that of standard single-model finetuning.

\paragraph{PCA} On certain datasets, applying G2T-FM directly on the original features results in out-of-memory errors. To address this, we apply principal component analysis (PCA) on \texttt{avazu-ctr}, \texttt{city-reviews}, \texttt{amazon-ratings} and \texttt{questions} datasets to reduce the feature dimensionality. PCA is performed separately on the original features and the neighborhood feature aggregations, reducing each to 64 dimensions.

\paragraph{PEARL and SF} For Laplacian positional encodings (LapPE), we set the dimension to $k = 14$ so that the total number of structure-based features (specifically, LapPE, degree, and PageRank) is $16$. For the GNN backbone within PEARL, we use a 3-layer GCN with GELU activations, hidden dimension $512$ and output dimension $16$. However, we do not use layer normalization and residual connections, based on preliminary experiments that showed improved results without these components. In PEARL, we average GNN embeddings over 8 random inputs.

For the in-context learning experiments, we utilize a randomly initialized PEARL model whose weights are shared across all datasets. Interestingly, even without explicit training, this untrained PEARL model still produces useful representations for some datasets, as demonstrated in our ablation studies (Section~\ref{sec:ablation}). For the finetuning experiments, we jointly finetune PEARL and the TFM backbone.

\subsection{GNNs and LightGBM}

\paragraph{GNNs} Our GNN setup closely follows the architecture and hyperparameter optimization procedure from GraphLand~\citep{GraphLand}, with two main differences. First, we introduce early stopping with a patience of 100 steps to accelerate training. Second, we use unified hyperparameter search spaces for both feature and target preprocessing, rather than dataset-specific spaces used in GraphLand. See our code for more details. Note that this latter modification only affects the preprocessing hyperparameters, while the search grids for learning rate and dropout remain identical to those in GraphLand.

\paragraph{PEARL integration} In the ablation studies described in Section~\ref{sec:ablation}, we evaluate the effect of integrating PEARL with both GNN and LightGBM models. For GNNs, we concatenate PEARL outputs with the initial node features, and train the combined model end-to-end. For LightGBM, due to the challenge of end-to-end training with PEARL, we use the outputs from the same randomly initialized PEARL as in our G2T-FM (ICL) experiments.

\section{Concurrent and follow-up works}

\begin{table*}[t!]
\caption{Comparison of downstream performance between G2T-FM and TAG.}
\label{tab:tag-eval}
\begin{center}
\resizebox{\textwidth}{!}{
\begin{tabular}{lcccccccc}
\toprule
 & \texttt{amazon-ratings} & \texttt{artnet-exp} & \texttt{city-reviews} & \texttt{facebook} & \texttt{pubmed} & \texttt{questions} & \texttt{tolokers-2} & \texttt{wiki-cs} \\
\midrule
LightGBM-NFA & $43.05 \pm 0.11$ & $46.13 \pm 0.04$ & $78.53 \pm 0.01$ & $88.48 \pm 0.07$ & $89.31 \pm 0.05$ & $14.85 \pm 0.23$ & $56.34 \pm 0.06$ & $80.43 \pm 0.06$ \\
GCN & $41.43 \pm 0.49$ & $44.86 \pm 0.36$ & $77.81 \pm 0.15$ & $91.26 \pm 0.21$ & $85.46 \pm 0.19$ & $15.42 \pm 0.67$ & $56.27 \pm 0.31$ & $81.74 \pm 0.21$ \\
GraphSAGE & $40.07 \pm 0.53$ & $45.14 \pm 0.36$ & $78.17 \pm 0.10$ & $91.12 \pm 0.22$ & $86.04 \pm 0.27$ & $16.55 \pm 0.64$ & $54.43 \pm 0.34$ & $81.50 \pm 0.27$ \\
GAT & $40.67 \pm 0.55$ & $45.06 \pm 0.52$ & $77.74 \pm 0.21$ & $92.61 \pm 0.21$ & $84.81 \pm 0.24$ & $16.75 \pm 0.67$ & $57.41 \pm 0.85$ & $82.25 \pm 0.27$ \\
GT & $41.56 \pm 0.40$ & $46.41 \pm 0.71$ & $77.34 \pm 0.21$ & $91.71 \pm 0.22$ & $84.95 \pm 0.19$ & $14.03 \pm 0.90$ & $56.98 \pm 0.55$ & $82.54 \pm 0.21$ \\
\midrule
TAG-TabPFNv2 (ICL) & $43.97 \pm 0.52$ & $47.87 \pm 0.39$ & $77.38 \pm 0.29$ & $93.11 \pm 0.17$ & $87.80 \pm 0.20$ & $15.07 \pm 1.32$ & $59.33 \pm 1.00$ & $82.59 \pm 0.12$ \\
G2T-TabPFNv2 (ICL) & $41.99 \pm 0.12$ & $45.73 \pm 0.03$ & $77.28 \pm 0.25$ & $91.27 \pm 0.12$ & $89.39 \pm 0.15$ & $18.46 \pm 0.58$ & $60.39 \pm 0.19$ & $80.09 \pm 0.26$ \\
\midrule
TAG-LimiX (ICL) & $44.89 \pm 0.15$ & $50.19 \pm 0.46$ & $78.87 \pm 0.13$ & $92.85 \pm 0.14$ & $88.09 \pm 0.18$ & $15.53 \pm 1.99$ & $59.46 \pm 1.05$ & $82.64 \pm 0.17$ \\
G2T-LimiX (ICL) & $45.03 \pm 0.10$ & $48.42 \pm 0.78$ & $78.98 \pm 0.44$ & $91.72 \pm 0.10$ & $89.25 \pm 0.14$ & $20.97 \pm 0.52$ & $61.60 \pm 0.18$ & $80.93 \pm 0.13$ \\
\bottomrule
\end{tabular}
}
\end{center}
\end{table*}

\begin{table*}[t!]
\caption{Runtime in seconds per single seed.}
\label{tab:tag-runtime}
\begin{center}
\resizebox{\textwidth}{!}{
\begin{tabular}{lcccccccc}
\toprule
 & \texttt{amazon-ratings} & \texttt{artnet-exp} & \texttt{city-reviews} & \texttt{facebook} & \texttt{pubmed} & \texttt{questions} & \texttt{tolokers-2} & \texttt{wiki-cs} \\
\midrule
G2T-TabPFNv2 (ICL) & $169.04$ & $153.41$ & $150.14$ & $56.75$ & $236.52$ & $481.67$ & $9.38$ & $57.66$ \\
TAG-TabPFNv2 (ICL) & $736.37$ & $1{,}407.83$ & $2{,}308.25$ & $656.29$ & $568.63$ & $1{,}414.72$ & $150.83$ & $291.5$ \\
G2T-LimiX (ICL) & $116.28$ & $103.24$ & $279.41$ & $48.82$ & $151.02$ & $317.75$ & $18.00$ & $60.05$ \\
TAG-LimiX (ICL) & $825.71$ & $1{,}532.35$ & $2{,}479.54$ & $686.38$ & $600.54$ & $1{,}539.22$ & $150.04$ & $310.86$ \\
\bottomrule
\end{tabular}
}
\end{center}
\end{table*}

Shortly after the release of this manuscript, several independent works explored applying TFMs and PFNs to graph node-level tasks. 
The closest concurrent methods are TAG~\citep{TAG} and TabPFN-GN~\citep{choi2025can}. Like G2T-FM, they augment original features with hand-crafted graph-aware features (e.g., neighborhood aggregations or node degrees) and subsequently apply a TFM. However, their specific feature sets and implementations differ. For completeness, we discuss the key differences and provide an empirical comparison with TAG.\footnote{We do not compare with TabPFN-GN because, at the time of writing, its code has not been publicly released.}

\paragraph{Key differences} Our work differs from~\citet{TAG,choi2025can} in the following key ways. First, we focus on real-world datasets from the recently introduced GraphLand benchmark~\citep{GraphLand} which contains a variety of classification and regression tasks covering more realistic problems compared to classic graph benchmarks that have been recently criticized in~\citet{bechler2025position}. Second, we evaluate TFMs in both in-context learning and finetuning regimes, and we show that finetuning substantially improves performance. Finally, we use well-tuned GNNs with hyperparameter tuning and architectural enhancements that serve as strong baselines for GFMs.

\paragraph{Evaluation} We evaluate TAG in our setup (i.e., on datasets and splits used in our main experiments) using the official implementation of TAG provided by the authors. The only change we make is to replace the accuracy metric with average precision during ensemble selection and evaluation on binary classification datasets, because average precision is the metric used throughout our study. We restrict the comparison to classification datasets because the current TAG implementation does not support regression tasks at the time of writing. Table~\ref{tab:tag-eval} reports the results. Overall, the two methods achieve comparable performance. However, our preliminary runtime benchmarks in Table~\ref{tab:tag-runtime} indicate that G2T-FM is typically 4-10 times faster.

\paragraph{Follow-up work} Other relevant works are GraphPFN~\citep{eremeev2026graphpfn} and NodePFN~\citep{NodePFN}. Building on the ideas and results of G2T-FM, GraphPFN replaces hand-crafted features with learnable message-passing layers embedded into the LimiX TFM. These layers are pretrained on synthetic datasets using the PFN framework. NodePFN adopts a similar approach but pretrains the entire model from scratch rather than initializing from an existing TFM. 

Importantly, while methods such as GraphPFN and NodePFN may yield more expressive graph-specific models, they lack G2T-FM's plug-and-play capability to easily incorporate any off-the-shelf TFM without additional pretraining. Given the rapid pace of TFM development in terms of both downstream quality and scalability~\citep{TabPFNv3, qu2026tabiclv2}, we believe this plug-and-play property is an important advantage of G2T-FM, allowing it to remain a strong and up-to-date baseline despite the presence of more advanced and complex approaches.


\end{document}